\documentclass{article}

\usepackage[final, nonatbib]{neurips_2023_gaied}  %

\usepackage{enumitem}
\usepackage[utf8]{inputenc} 
\usepackage[T1]{fontenc}    
\usepackage{hyperref}       
\usepackage{url}            
\usepackage{booktabs}       
\usepackage{amsfonts}       
\usepackage{nicefrac}       
\usepackage{microtype}      
\usepackage{xcolor}         

\usepackage{graphicx}

\title{Ruffle\&Riley: Towards the Automated Induction of Conversational Tutoring Systems}

\author{%
  Robin Schmucker \\
  Machine Learning Department\\
  Carnegie Mellon University\\
  \texttt{rschmuck@cs.cmu.edu} \\
  \And
  Meng Xia \\
  Human-Computer Interaction Institute\\
  Carnegie Mellon University\\
  \texttt{mengxia@cs.cmu.edu}\\
  \And
  Amos Azaria\\
  Department of Computer Science\\
  Ariel University\\
  \texttt{amos.azaria@ariel.ac.il} \\
  \And
  Tom Mitchell\\
  Machine Learning Department\\
  Carnegie Mellon University\\
  \texttt{tom.mitchell@cs.cmu.edu} \\
}

\begin{document}

\maketitle

\begin{abstract}
Conversational tutoring systems (CTSs) offer learning experiences driven by natural language interaction. They are known to promote high levels of cognitive engagement and benefit learning outcomes, particularly in reasoning tasks. Nonetheless, the time and cost required to author CTS content is a major obstacle to widespread adoption. In this paper, we introduce a novel type of CTS that leverages the recent advances in large language models (LLMs) in two ways: First, the system induces a tutoring script automatically from a lesson text. Second, the system automates the script orchestration via two LLM-based agents (Ruffle\&Riley) with the roles of a student and a professor in a learning-by-teaching format. The system allows a free-form conversation that follows the ITS-typical inner and outer loop structure. In an initial between-subject online user study (N = 100) comparing Ruffle\&Riley to simpler QA chatbots and reading activity, we found no significant differences in post-test scores. Nonetheless, in the learning experience survey, Ruffle\&Riley users expressed higher ratings of understanding and remembering and further perceived the offered support as more helpful and the conversation as coherent. Our study provides insights for a new generation of scalable CTS technologies.
\end{abstract}

\section{Introduction}
\label{sec:introduction}

Intelligent tutoring systems (ITSs) are a type of educational technology that provides millions of learners worldwide with access to learning materials and affordable personalized instruction. ITSs can, in certain cases, be as effective as human tutors~\cite{Vanlehn2011:Relative, Kulik2016:Effectiveness} and can play an important role in mitigating the educational achievement gap~\cite{Roschelle2016:Online, Huang2016:Intelligent}. However, despite their potential, one major obstacle to the widespread adoption of ITS technologies is the large costs associated with content development. Depending on the depth of instructional design and available authoring tools, preparing one hour of ITS content can take instructional designers hundreds of hours~\cite{Aleven2016:Example}.

Conversational tutoring systems (CTSs) are a type of ITS that engages with learners in natural language. Various studies have confirmed the benefits of CTSs, across multiple domains, particularly on learning outcomes in reasoning tasks~\cite{Paladines2020:Systematic}. Still, many existing CTSs struggle to maintain coherent free-form conversations and understand the learners' responses due to limitations imposed by their underlying natural language processing (NLP) techniques~\cite{Graesser2012:Autotutor}. In this paper, we introduce a new type of CTS that draws inspiration from design principles of earlier CTSs~\cite{Nye2014:Autotutor, leelawong2008designing} and that leverages the recent advances in large language models (LLMs)~\cite{Zhou2023:Comprehensive} to automate content authoring and to orchestrate free-form conversational tutoring.
Our main contributions include:
\begin{itemize}
[itemsep=3.0pt,topsep=0.0pt,leftmargin=*]
    \item \textit{Automated Induction of a Conversational Tutoring System from Text:} We introduce a new type of CTS that employs LLMs to generate a tutoring script automatically from a lesson text and that further automates the script orchestration in a free-form conversation. In particular, we orchestrate the conversation in a learning-by-teaching format via two conversational agents taking on the roles of a student (Ruffle) and a professor (Riley). The conversation follows the prototypical ITS structure by exhibiting an outer loop (problem sequencing) and an inner loop (feedback/assistance)~\cite{Vanlehn2006:Behavior}.
    \item \textit{Findings from an Online User Study}
    We report findings from a user study that evaluates the effects of our LLM-induced CTS workflow on learning outcomes and user experience, comparing it to two simpler QA chatbots and reading activity. We discuss the strengths and limitations of our current system, describe our plans for system refinements in response to user feedback, and provide guidance for the design and evaluation of future LLM-based CTSs.
\end{itemize}

\begin{figure}[t]
    \centering
    \includegraphics[width=1.0\textwidth]{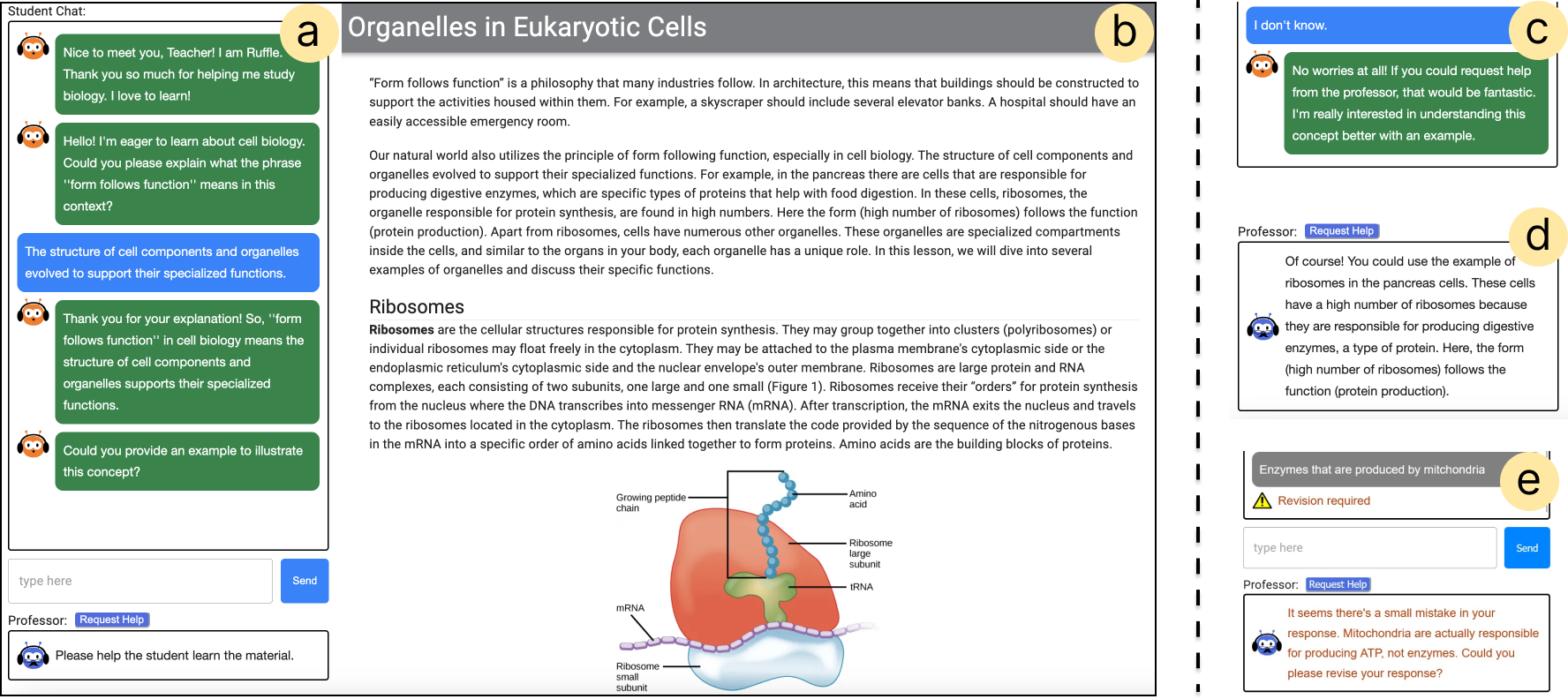}
    \caption{UI of Ruffle\&Riley. (a) Learners are asked to teach Ruffle (student agent) in a free-form conversation and request help as needed from Riley (professor agent). Ruffle tries to guide the learner to articulate the expectations in the tutoring script. (b) The learner can navigate the lesson material during the conversation. (c) Ruffle encourages the learner to explain the content. (d) Riley responds to a help request. (e) Riley detected a misconception and prompts the learner to revise their response.}
    \label{fig:ui}
\end{figure}

\section{Related Work}
\label{sec:related_work}

\subsection{Conversational Tutoring Systems}

Dialog-based learning activities are known to lead to high levels of cognitive engagement~\cite{Chi2014:Icap}, and various studies have confirmed their benefits on learning outcomes (e.g.,~\cite{Cohen1982:Educational, Chi2001:Learning}). This motivated the integration of conversational activities into learning technologies. In their systematic review, Paladines and Ramirez~\cite{Paladines2020:Systematic} categorized the design principles underlying existing CTSs into three major categories: (i) expectation misconception tailoring (EMT)~\cite{D2012:Gaze, Nye2014:Autotutor}, (ii) model-tracing (MT)~\cite{Rose2001:Interactive, Rickel2002:Collaborative, Heffernan2008:Expanding}) and (iii) constraint-based modeling (CBM)~\cite{Mitrovic2005:Effect, Weerasinghe2006:Facilitating}. While all three frameworks can promote learning, they require instructional designers to spend substantial effort configuring the systems for each individual lesson and domain. Further, due to limitations of underlying NLP techniques, many CTSs struggle to maintain coherent free-form conversations, answer learners' questions, and understand learners' responses reliably~\cite{Nye2014:Autotutor}. In this context, this paper employs recent NLP advances as the foundation for a novel type of LLM-driven CTSs that can orchestrate coherent free-form adaptive dialogues and that can alleviate the burdens associated with content authoring.

\subsection{Content Authoring Tools}

One major obstacle to the widespread adoption of CTSs and other types of ITSs is the complexity and cost of content authoring~\cite{Aleven2006:Cognitive, Graesser2012:Autotutor, Dermeval2018:Authoring}. For early ITSs, the development ratio (i.e., the number of hours required to author one hour of instructional content) was estimated to vary between 200:1 and 300:1~\cite{Aleven2006:Cognitive}. Content authoring tools (CATs)~\cite{Murray2003:Overview} were developed to facilitate ITS creation, often with an emphasis on making the process accessible to educators without programming background (e.g.,~\cite{Koedinger2004:Opening, Wolfe2013:Development, Aleven2016:Example}). While a comprehensive survey of CATs is vastly beyond the scope of this paper--for this, we refer to~\cite{Dermeval2018:Authoring, Sottilare2015:Design}--here we focus on highlighting prior studies that illustrate the ability of existing CATs to reduce authoring times. ASSISTment Builder~\cite{Razzaq2009:Assistment} was developed to support content authoring in a math ITS and enabled a development ratio of 40:1. For model tracing-based ITSs, example tracing~\cite{Aleven2016:Example} has proven itself as an effective authoring technique that depending on the context enables development ratios between 50:1 and 100:1. Recently, apprentice learner models were evaluated as another authoring technique that in certain cases can be more efficient than example tracing~\cite{Weitekamp2020:Interaction, Maclellan2022:Domain}. In the context of CTSs, multiple CATs have been developed for AutoTutor~\cite{Cai2019:Authoring}, and while we were not able to find concrete development ratio estimates, the authoring of CTS content 
is still considered to be complex and labor intensive.

Alternative approaches explored the use of learner log data to enhance ITS components such as skill models and hints (e.g., \cite{Barnes2005:Q, Cen2006:Learning, Barnes2008:Toward}) as well as machine learning-based techniques for automated questions and feedback generation (e.g.,~\cite{Kurdi2020:Systematic, Hahn2021:Systematic}). Recent advances in large language models (LLMs)~\cite{Zhou2023:Comprehensive} sparked a new wave of research that explores ways in which LLM-based technologies can benefit learners~\cite{Kasneci2023:Chatgpt}. Settings in which LLMs already have been found to be effective include question generation and quality assessment~\cite{Ruan2019:Bookbuddy, Huy2022:Towards, Moore2023:Assessing, Jiao2023:Automatic, Ahmed2023:Chatgpt}, feedback generation~\cite{Ruan2019:Bookbuddy, Jin2022:Stubot, Nguyen2023:Evaluating, Liffiton2023:Codehelp, Pardos2023:Learning, Zografos2023:Gpt, Ahmed2023:Chatgpt}, answering students' questions~\cite{Lee2023:Dapie, Sonkar2023:Class}, automated grading~\cite{Hirunyasiri2023:Comparative, Botelho2023:Leveraging}, and helping teachers reflect on their teaching~\cite{Markel2023:Gpteach, Lin2023:Using, Demszky2023:Can}. What sets this paper apart from the aforementioned works is that it does not focus on the generation of \textit{individual} ITS components; instead, we propose a system that can automatically induce a \textit{complete ITS workflow}, exhibiting the prototypical inner and outer loop structure~\cite{Vanlehn2006:Behavior}, directly from a lesson text. Our work represents a step towards LLM-driven ITS authoring tools that can generate entire workflows automatically from existing learning materials and reduce system development times by an order of magnitude potentially.

\section{System Architecture}
\label{sec:system_overview}

\textbf{Design Considerations}

We approached the design of Ruffle\&Riley with two specific goals in mind: (i) Facilitate an ITS workflow that provides learners with a sequence of questions (outer loop) and meaningful feedback during problem-solving (inner loop); (ii) Streamline the process of configuring the conversational agents for different lesson materials. We reviewed existing CTSs and identified EMT~\cite{Graesser2004:Autotutor} as a design framework suitable for our objectives. EMT mimics teaching strategies employed by human tutors~\cite{Graesser1995:Collaborative} by associating each question with a list of expectations and anticipated misconceptions. After presenting a question and receiving an initial user response, EMT-based CTSs provide inner loop support (goal (i)) by guiding the conversation via a range of dialogue moves to correct misconceptions and to help the learner articulate the expectations before moving to the next question (outer loop). While EMT-based CTSs have been shown to be effective in various domains~\cite{Nye2014:Autotutor}, they need to be configured in a labor-intensive process that requires instructional designers to define a \textit{tutoring script} that specifies questions, expectations, misconceptions and other information for each lesson~\cite{Cai2019:Authoring}. For us, tutoring scripts are attractive as a standardized format for CTS configuration (goal (ii)).

\begin{figure}[t]
    \centering
    \includegraphics[width=0.85\textwidth]{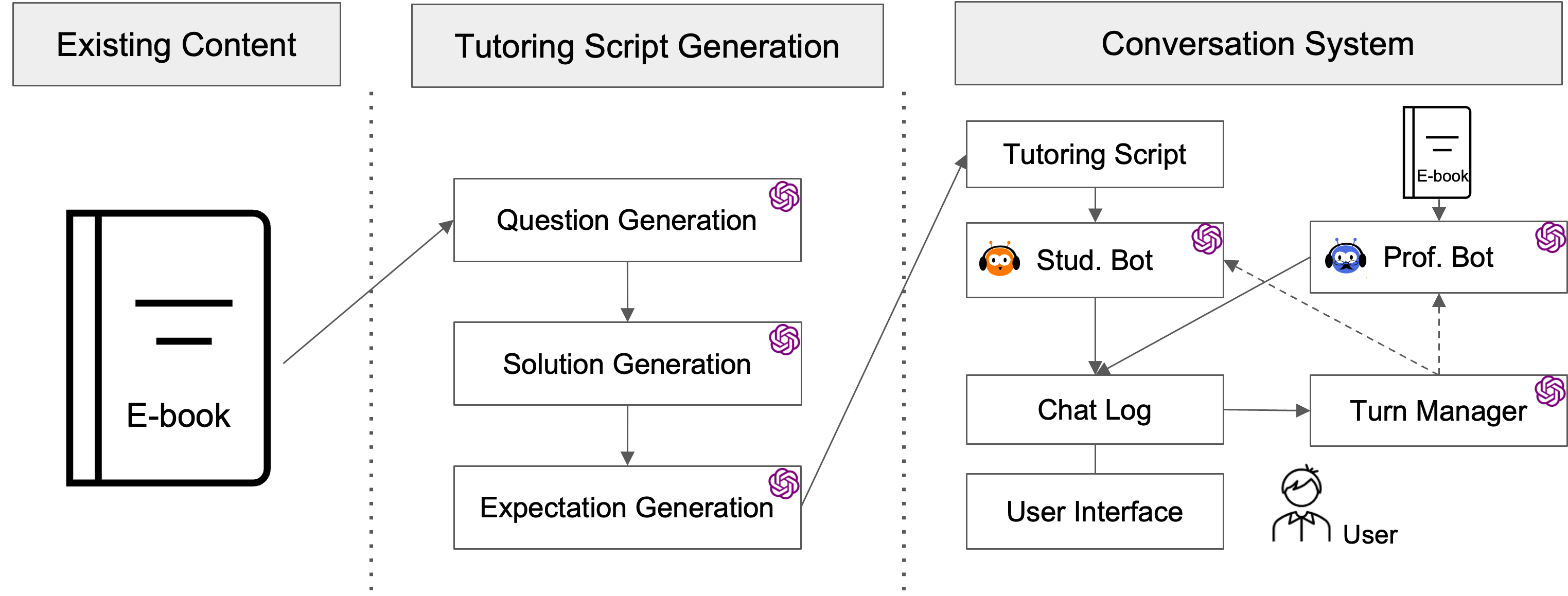}
    \caption{System architecture. Ruffle\&Riley generates a \textit{tutoring script} automatically from a lesson text by executing three separate prompts that induce questions, solutions and expectations for the EMT-based dialog. During the learning process, the script is orchestrated via two LLM-based conversational agents in a free-form dialog that follows the ITS-typical outer and inner loop structure.}
    \label{fig:system}
\end{figure}

An overview of our user interface, together with descriptions of its key elements, is provided by Figure~\ref{fig:ui}. Inspired by the success of learning-by-teaching activities~\cite{duran2017learning, fiorella2013relative, leelawong2008designing}, we decided to orchestrate the conversation in a learning-by-teaching format via two conversational agents taking on the roles of a student (Ruffle) and a professor (Riley). While our design is similar to some CTSs in the AutoTutor family~\cite{Nye2014:Autotutor} that follow a trialogue format, one notable difference is that Riley solely serves as an assistant to the learner by offering assistance and correcting misconceptions. Riley never communicates with Ruffle directly. In the following, we describe the system architecture underlying Ruffle\&Riley in more detail (Figure~\ref{fig:system}).

\textbf{Tutoring Script Generation} Ruffle\&Riley is capable of generating a tutoring script fully automatically from a lesson text by leveraging GPT4~\cite{Openai2023:Gpt4}. This involves a 4-step process: (i) A list of review questions is generated from the lesson text; (ii) For each question, a solution is generated based on question and lesson texts; (iii) For each question, a list of expectations is generated based on question and solution texts; (iv) The final tutoring script is compiled as a list of questions together with related expectations. The first three steps are implemented via three separate prompts written in a way general enough to support a wide range of lesson materials. Unlike traditional EMT-based CTSs, our tutoring scripts do not attempt to anticipate misconceptions learners might exhibit ahead of time (this is a difficult task even for human domain experts). Instead, we rely on GPT4's ability to detect factually incorrect information in the learner's responses during the active teaching process.

\textbf{Conversation Orchestration} EMT-based CTSs require the definition of dialog moves and conversational turn management to facilitate coherent conversations, which in itself is a complex authoring process~\cite{Cai2019:Authoring}. Ruffle\&Riley automates the tutoring script orchestration by including descriptions of desirable properties of EMT-based conversations into the agents' prompts and captures the user's state solely via the chat log. The student agent receives the tutoring script as part of its prompt and is instructed to let the user explain the individual questions and to ask follow-ups until all expectations are covered. Ruffle reflects on user responses to show understanding, provides encouragement to the user, and keeps the conversation on topic. In parallel, Riley's prompt contains the lesson text and instructions to provide feedback to user responses, to offer relevant information after help requests, and to prompt the user to revise their response after detecting incorrect information. Both agents are instructed to keep the conversation positive and encouraging and to not refer to information outside the tutoring script/lesson text. The turn manager coordinates the system's queries to GPT4.

\section{System Evaluation}
\label{sec:evaluation}

\subsection{Experimental Design}
\label{subsec:experiment_design}
The recruitment and study process was approved by the Institutional Review Board (IRB).

\textbf{Content} To evaluate the efficacy of the system, we adapted a Biology lesson on organelles of eukaryotic cells from the OpenStax project~\cite{Biology2e:OpenStax}. We selected this lesson because we expected participants to have low prior familiarity with the material to ensure a learning process. Nevertheless, the lesson text is designed to be accessible to a general audience.

\textbf{Conditions} Similar to prior work~\cite{Vanlehn2011:Relative, Kopp2012:Improving}, we construct conditions to compare the efficacy of our EMT-based CTS to reading alone and to a QA chatbot with limited dialog. To study potential differences, we equip the QA chatbot with content from different sources under two distinct conditions: one using content generated by a biology teacher and the other using content from the LLM.

\begin{itemize}[itemsep=0.5pt,topsep=0.0pt,leftmargin=*]
    \item Reading: Participants study the material without receiving additional support.
    \item Teacher QA (TQA): Participants study the material and can answer review questions presented by the chatbot. After submitting an answer, participants receive brief feedback about the correctness of their answer and a sample solution. Questions and answers are designed by a teacher.
    \item LLM QA (LQA): Same as TQA, but questions and answers are generated by the LLM (Section~\ref{sec:system_overview}).
    \item Ruffle\&Riley (R\&R): Participants study the material while being supported by the two conversational agents (Section~\ref{sec:system_overview}). The system is equipped with the same questions as LQA.
\end{itemize}

\textbf{Surveys/Questionnaires}
We evaluate system efficacy from two perspectives: \textit{learning performance} and \textit{learning experience}. Performance is captured via a multiple-choice post-test after the learning session, which consists of five questions written by a biology teacher recruited via Upwork~\cite{Upwork:2023} and two questions from OpenStax~\cite{Biology2e:OpenStax}. 
The learning experience is captured via a 7-point Likert scale questionnaire that queries participants' perception of engagement, intrusiveness, and helpfulness of the agents, based on previous work~\cite{peng2022crebot}. To ensure data quality, we employed two attention checks and one question asking participants whether they looked up test answers online. Further, we included a demographic questionnaire to understand participants' age, gender, and educational background.

\textbf{Participants} 
We recruited participants located in the USA who were fluent in English and had at least a high-school (HS) degree via Prolific~\cite{Prolific:2023}.
Participants were randomly assigned to the conditions and were free to drop out at any point in the study. Overall, 100 participants completed the task. As shown in Table~\ref{fig:performance}, 30 participants finished the reading condition, 17 finished TQA, 23 finished LQA, and 30 finished R\&R. The imbalance is due to random condition assignments and dropouts.

\textbf{Hypotheses} We explore the following hypotheses. H1: \textit{Learning Outcomes}: R\&R achieves higher post-test scores than the baseline conditions (H1a); There is no significant difference between TQA and LQA (H1b). H2: \textit{Learning experience}: R\&R achieves higher ratings than the baseline conditions in terms of engagement, helpfulness in understanding, remembering, interruption, coherence, support received, and enjoyment (H2a); There are no significant differences between TQA and LQA (H2b).

\subsection{Results and Analysis}
\label{subsec:results}

After filtering participants who failed any of the attention check questions, or who did not rate ``strongly disagree'' when asked whether they looked up test answers, we were left with 58 (male: 33, female: 21, other: 4) out of the 100 participants (15 in reading, 7 in TQA, 15 in LQA, and 21 in R\&R). The age distribution is 18-25 (8), 26-35 (20), 36-45 (18), 46-55 (9), over 55 (3). The degree distribution is: HS or Equiv. (22) Bachelor's/Prof. Degree (25), Master's or Higher (11).

\textbf{Learning Performance} The post-test consists of seven questions, each worth one point. The mean and standard error in post-test scores for 
each condition is provided by Table~\ref{fig:performance}. A one-way ANOVA did not detect significant differences in post-test scores among the four conditions. Therefore, we find support for H1b but not for H1a. Even though not significantly different, we observed that participants in R\&R achieved somewhat higher scores ($5.19 \pm 0.25$) than in TQA ($4.14 \pm 0.83$).

\begin{table}[t]
    \centering
     \caption{Learning performance across different conditions.} \includegraphics[width=1.0\textwidth]{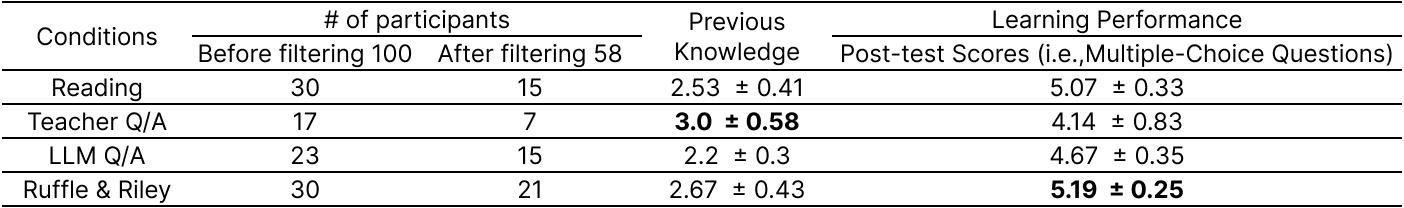}
    \label{fig:performance}
\end{table}

\begin{table}[t]
    \centering
    \caption{Learning experience across different conditions. The symbol "*" represents $p < 0.05$. The symbol "-" represents that this aspect was not asked in the corresponding condition.}
    \includegraphics[width=1.0\textwidth]{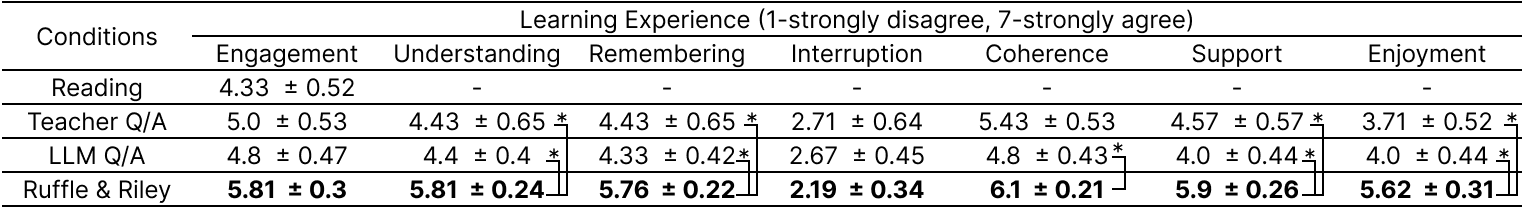}
    \label{fig:experience}
\end{table}
\textbf{Learning Experience} Table~\ref{fig:experience} shows participants' learning experience and chatbot interaction ratings. We tested for significance ($p < 0.05$) via one-way ANOVA, followed by Bonferroni post-hoc analysis. We found no significant differences in self-reported engagement levels between the four conditions. However, among the three chatbot conditions, R\&R was rated as significantly more helpful in aiding participants in understanding, remembering the lesson and providing the support needed to learn. Further, R\&R participants expressed more enjoyment than TQA and LQA participants. In addition, participants found R\&R provided a significantly more coherent conversation than LQA. Interestingly, even though we expected R\&R to be rated as more interrupting, we found no significant differences in perceived interruption among the three chatbot conditions. Therefore, H2a is partially supported. In addition, there were no significant differences detected between LQA and TQA among the aspects of the learning experience. Thus, we cannot reject H2b.

\textbf{{Evaluation of Conversations in R\&R}} We analyzed chat and behavior log data in R\&R. First, based on logged responses and scrolling events, we found that participants taught Ruffle with different strategies. Some participants learned the lesson driven by the questions asked by Ruffle (10 out of 21) and focused on the conversation immediately after entering the learning session before reading the whole lesson text. Other participants read the text first before starting the conversation. Second, on average, participants submitted $1.33 \pm 1.74$ help requests. Third, the mean learning time in each condition is: reading (4 mins), TQA (11 mins), LQA (12 mins), and R\&R (18 mins).

\section{Discussion and Limitations}
\label{sec:discussion}

Here, we discuss lessons learned, future directions, and limitations.

\textbf{Extending the System Evaluation} While our evaluation showed that Ruffle\&Riley can improve various learning experience metrics, we were not able to detect significant improvements in post-test scores. Our post-test featured recall-based multiple-choice questions, which represent a shallow form of knowledge assessment. Our observations align with prior research that found that reading and conversational tutoring can lead to similar outcomes in recall-based test formats~\cite{Graesser2004:Autotutor}. As a next step, we want to revise our post-test format and employ fill-in-the-blank and essay questions to assess deeper understanding. We also want to evaluate knowledge retention over time and gauge the system's ability to facilitate learning in other domains (e.g., psychology and business).

\textbf{Refining the Instructional Design} Some users engaged in the conversational workflow before reading the lesson text and focused exclusively on the questions presented by the agents. This can cause users to miss important information that falls outside the scope of the tutoring script. In future evaluations, we want to orchestrate the learning activity in a more structured way, either by requiring the users to read the lesson text first or by interweaving lesson material with conversation~\cite{Winkler2020:Sara}. Further, we observed that conversational learning activities require substantially more time than reading. This motivates us to explore more time-efficient CTS workflows~\cite{Kopp2012:Improving} in future research.

\textbf{Human-in-the-loop Capabilities} 
The present study leverages an LLM to automatically generate a tutoring script from a lesson text and automates its orchestration in a free-form dialog via two conversational agents. While our study showcases the impressive capabilities of state-of-the-art NLP technologies, we think that it is crucial to move towards extending the system architecture with human-in-the-loop capabilities~\cite{Xia2019:Peerlens, Tong2023:Trustworthy}. We want to enable general educational practitioners to get involved in the instruction design process. For example, teachers might want to include their own review questions or insert expectations for students' answers into the tutoring script or alternatively choose among different candidates that the LLM can provide to them. Including human domain experts in the design process can further improve trustworthiness and content quality.

\textbf{Limitations} Ruffle\&Riley is still in an early stage of development, and the present study is subject to several limitations. First, the system was evaluated in an online user study conducted via Prolific~\cite{Prolific:2023} with adult participants exhibiting diverse demographics (i.e., age and education). Current findings focus on a broad population of online users and might not generalize to more specific populations (e.g., K12 or college students). Before evaluating Ruffle\&Riley with younger learners in institutional settings, we need to certify safe and trustworthy system behavior~\cite{Escobar2022:Guidelines}. Relatedly, we need to verify the factual correctness of the information that surfaces during the conversations~\cite{Kasneci2023:Chatgpt}. While we instructed the GPT4-based agents to only refer to information that occurs in the lesson text, and we have not observed incorrect information during system testing and data analysis, a systematic evaluation is fundamental. A related direction is the integration of safeguards and other validations mechanisms into the system to ensure benign outputs. Another limitation is that participants only took part in a single learning session. Users might require some time to get used to the workflow.

\section{Conclusion}
\label{sec:conclusion}

In 2019, the team around AutoTutor~\cite{Cai2019:Authoring} reflected on the labor-intensive process in which conversational tutoring systems are created and hinted towards the possibility that, sometime in the future, one might be able to generate tutoring scripts fully automatically. Now, in 2023, we have reached a point where the capabilities of available NLP technologies enable us to make this vision become a reality. In the coming years, generative-AI-based content authoring tools are likely to allow researchers and educators to focus their limited time more on questions of effective instructional design and ITS architecture and less on system implementation. This might accelerate the evaluation of instructional design principles~\cite{Koedinger2013:Instructional}, leading to improvements in ITSs and to general insights for learning science.

\begin{ack}
We would like to thank Art Graesser for helpful suggestions and critiques during system development. We thank Xiaoyang Lei for creating illustrations for our two agents. We further thank Microsoft for support in the form of Azure computing and access to the OpenAI API through a grant from their Accelerate Foundation Model Academic Research Program. This research was supported by the AFOSR under award FA95501710218.
\end{ack}

\medskip
\bibliographystyle{plain}
\bibliography{bibliography}


\end{document}